\documentclass[]{spie}  

\usepackage{amsmath,amsfonts,amssymb}
\usepackage{graphicx}
\usepackage[colorlinks=true, allcolors=blue]{hyperref}
\usepackage[T1]{fontenc}
\usepackage{microtype}
\hypersetup{
  pdftitle={Toward surface-based registration of a virtual preoperative cutting guide onto the mandible for reconstruction surgery},
  pdfauthor={Yue Yang, Jie Ying Wu},
  pdfkeywords={augmented reality, mandible reconstruction, cutting guide, markerless registration, teeth}
}

\title{Toward surface-based registration of a virtual preoperative cutting guide onto the mandible for reconstruction surgery}

\author[a]{Yue Yang}
\author[a]{Jie Ying Wu}
\affil[a]{Vanderbilt University, Department of Computer Science, 1400 18th Avenue South, Suite A4004 Nashville, TN 37212, USA.}

\authorinfo{Further author information: (Send correspondence to Jie Ying Wu)\\Jie Ying Wu: E-mail: jieying.wu@vanderbilt.edu}

\begin{document} 
\maketitle

\begin{abstract}
Mandibular reconstruction restores facial continuity and oral function after segmental resection. Patient-specific cutting guides transfer a computed tomography (CT)-based plan to the operating room with three-dimensional information, but printed guides add cost and lead time, cannot adapt after fabrication, and may interrupt surgery if sterility is lost. We investigate a markerless augmented reality (AR) alternative that registers a virtual cutting guide to the exposed mandible from surface geometry. The method extends surface-based registration for the transoral setting, where teeth form the most distinctive visible surface. After camera calibration, a HoloLens 2 time-of-flight camera captures a partial intraoperative point cloud. The user supplies a rough head-based alignment only to crop the region of interest. A teeth-weighted global stage computes correspondences and solves a truncated least-squares rigid alignment. An asymmetric point-to-plane iterative closest point (ICP) stage refines the complete CT mandible against the partial depth cloud target. The guide-to-mandible transform places the guide in the HoloLens world frame, while pose updates and interpolation follow target motion. We define a blinded phantom protocol with 30 target registration error (TRE) points under full, intermediate, and teeth-only exposure, plus a motion-to-display latency test. Our median TRE is 4.05, 6.10, and 7.10~mm respectively, and median latency is 0.805~s. These values support the feasibility of using AR to replace physical prints. The workflow removes mounted fiducials and manual landmark selection and provides a testable path toward transoral AR cutting guidance.
\end{abstract}


\section{INTRODUCTION}
Segmental mandibular defects arise after oncologic resection, trauma, infection, or osteoradionecrosis (bone death). Reconstruction must restore a stable arch for mastication and speech while preserving the airway and facial contour. Fibula free-flap reconstruction remains common. One 21-year institutional series reported 413 flaps, including computer-aided design and manufacturing in only 40.7\% of cases.\cite{Allen2026}. Accessibility and costs could be significant contributing factors to this lack of guidance. Specifically, virtual surgical planning reconstructs the mandible from CT, defines resection and donor-bone osteotomies, and transfers those planes with patient-specific cutting guides. These workflows can reproduce the plan accurately and support complex multi-segment reconstruction.\cite{Annino2022}

A physical guide, however, must be designed, manufactured, shipped, sterilized, and seated on the intended anatomy. This process adds cost and time, and the printed guide cannot reflect an intraoperative change in margin or geometry.\cite{Ury2025,Coppen2025} Loss of sterility can also interrupt the workflow because a replacement cannot be fabricated during the operation. AR can instead project a modifiable virtual guide directly onto the operative field. Prior studies have shown virtual osteotomy guidance on fibular models, markerless tracking of a mandible, and markerless maxillary resection guidance.\cite{Coppen2025,Ury2025,Ceccariglia2022} Many systems still use attached fiducials, depend on learned views of a largely intact target, or lose reliability when tissue and instruments occlude the anatomy.\cite{Ury2025,Ceccariglia2022} The difficult transoral case is even more difficult with limited mandible exposure, and current systems face significant challenges. Specifically, the intraoperative camera capture may contain mostly teeth, while the preoperative source is the complete CT mandible.

We previously introduced a depth-based registration framework \cite{yang2026easyreg, yang2026all} and adapted it here to use teeth as automatically selected natural landmarks. Our contributions are: 1) The method registers the guide using teeth as the primary registration target without a marker fixed to bone and without manual landmark selection; 2) The overlay pose updates after rigid target motion, with a small delay that we quantify; 3) The evaluation protocol controls three exposure levels that approximate increasing surgical occlusion.

\section{METHOD}
\subsection{Workflow and coordinate transformations}
Preoperative CT segmentation produces a complete mandibular surface point set $P_M=\{p_i\in\mathbb{R}^3\}$ in mandible frame $\mathcal F_M$. Planning also defines the guide mesh in frame $\mathcal F_G$ and the fixed transform ${}^{M}T_G$. Before surgery, intrinsic calibration, depth-to-display calibration, and a device-specific working-distance bias correction determine the relation between the HoloLens~2 Articulated HAndTracking (AHAT) depth camera frame $\mathcal F_D$ and world frame $\mathcal F_W$. We use ${}^{A}T_B\in SE(3)$ to map a point from $\mathcal F_B$ to $\mathcal F_A$.

\begin{figure}[t]
\centering
\includegraphics[width=0.98\textwidth]{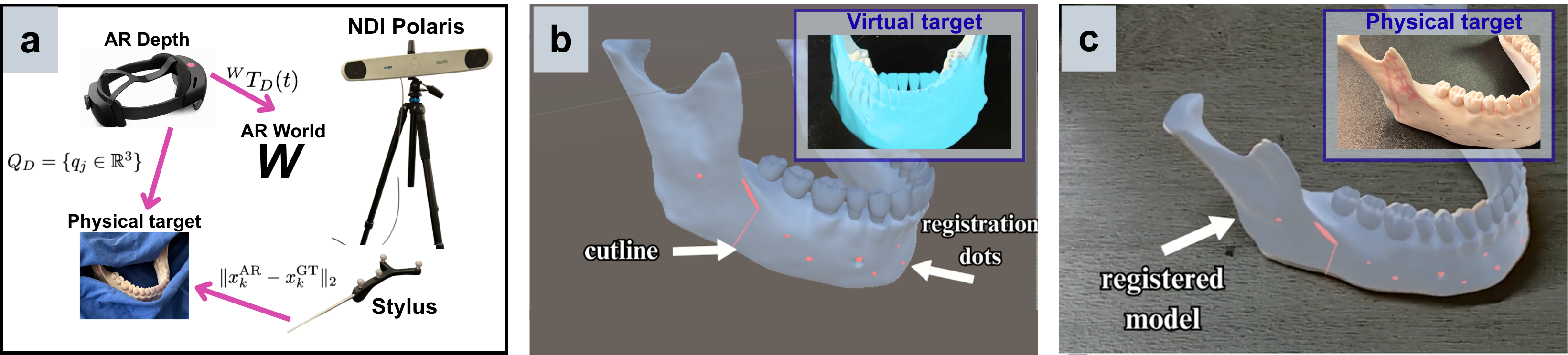}
\caption{Coordinate transform and system workflow. (a) Transformation between different components. Stylus and NDI Polaris are used for evaluation experiments only. (b) The pre-planed virtual mandible with registration target dots defined and virtual target visualized on AR during initial three-dimensional cropping. (c) The AR overlay showing the registered virtual model on top of the physical target under full exposure.}
\label{fig:workflow}
\end{figure}

Intraoperatively, the AHAT sensor captures a depth point set $Q_D=\{q_j\in\mathbb{R}^3\}$ that contains teeth, nearby exposed bone, soft tissue, and outliers (Figure~\ref{fig:workflow}a). A floating mandible target prompts the user to move their head until the virtual content in blue lies near the physical jaw. This interaction provides a coarse pose and a three-dimensional crop by removing most background points (Figure~\ref{fig:workflow}b). Registration estimates ${}^{D}\widehat T_M$ from the full CT surface to the partial depth surface. The displayed guide pose at time $t$ is
\begin{equation}
{}^{W}T_G(t) = {}^{W}T_D(t)\,{}^{D}\widehat T_M(t)\,{}^{M}T_G .
\label{eq:chain}
\end{equation}

\subsection{Teeth-aware global alignment and local ICP}
We applied a coarse-to-fine ICP structure but changed both stages for a small dental target.\cite{yang2026easyreg, yang2026markerless} We downsample $P_M$ and $Q_D$, estimate normals, and compute local covariance eigenvalues $\lambda_{1,i}\geq\lambda_{2,i}\geq\lambda_{3,i}$. Multi-scale surface variation
\begin{equation}
s_i=\max_{r\in\mathcal R}\frac{\lambda_{3,i}^{(r)}}{\lambda_{1,i}^{(r)}+\lambda_{2,i}^{(r)}+\lambda_{3,i}^{(r)}}
\label{eq:saliency}
\end{equation}
emphasizes cusp tips, incisal edges, or interproximal grooves that are more distinctive than broad mandibular surfaces. We then compute Fast Point Feature Histograms on both clouds and keep mutual nearest-neighbor feature matches $\mathcal C$. A match weight combines geometric saliency and descriptor similarity,
\begin{equation}
w_{ij}=(1+\gamma s_i)(1+\gamma s_j)
\exp\!\left(-\frac{\|\phi(p_i)-\phi(q_j)\|_2^2}{\sigma_\phi^2}\right).
\end{equation}
Because CT and depth coordinates share metric scale, the global stage fixes scale and uses the TEASER++ truncated least-squares formulation\cite{Yang2021TEASER}
\begin{equation}
(\widehat R_0,\widehat t_0)=\arg\min_{R\in SO(3),t}
\sum_{(i,j)\in\mathcal C}w_{ij}\min\!\left(\|q_j-(Rp_i+t)\|_2^2,c_g^2\right).
\label{eq:global}
\end{equation}
Curvature weighting increases the probability that the solver retains dental rather than smooth-body correspondences when the observed cloud contains little non-dental bone. 

The local stage must not force hidden parts of the complete CT model onto tissue or missing data. We therefore use an asymmetric scene-to-model objective. For each observed point $q_j$, we find the closest transformed model point $p_{i(j)}$ and retain it only when distance and normal-angle gates pass. With $T\leftarrow\exp(\xi^\wedge)T$, the update solves
\begin{equation}
\widehat\xi=\arg\min_{\xi}\sum_{j\in\mathcal I_\tau}
(1+\eta s_{i(j)})\,\rho_\sigma\!\left(
 n_{i(j)}^{\mathsf T}\left[q_j-\exp(\xi^\wedge)T p_{i(j)}\right]\right),
\label{eq:local}
\end{equation}
where $\rho_\sigma(r)=1-\exp[-r^2/(2\sigma^2)]$ is a Welsch loss. This robust point-to-plane form follows modern ICP practice.\cite{Zhang2022ICP} At each scale, $\mathcal I_\tau$ also rejects normal disagreement above $45^\circ$ and trims the largest 30\% of remaining residuals. We refine from 4 to 2 ~mm voxels, tighten the distance gate from 10 to 6 to 3~mm, and stop when the pose increment or cost change falls below the threshold. After initial registration, fast robust ICP uses the last accepted pose to crop the next depth frame and estimates ${}^{D}\widehat T_M(t)=\Delta T_t{}^{D}\widehat T_M(t-1)$. The system rejects an update when correspondence support is low or the pose jump exceeds our motion bound. The renderer interpolates between accepted poses. Interpolation smooths the overlay but does not reduce true motion-to-display delay.

We designed a phantom protocol around a PLA mandible printed on a Bambu Lab H2S to quantify the error of our method. Based on the pre-planned cutting guide on the virtual model, the mandible contained three cutting lines with start, mid, and end points defined on each line. We also scattered 21 virtual points on the mandible surface, resulting in 30 registration points per mandible. The points are printed black on the physical phantom and red on the AR overlay. We used an NDI Polaris camera to track a calibrated stylus (Figure~\ref{fig:workflow}a). Based on the registered model (Figure~\ref{fig:workflow}c), the experimenter placed the stylus tip at each virtual red target shown on the AR overlay ($x^{\mathrm{AR}}$) while the ground truth physical black-ink targets ($x^{\mathrm{GT}}$) were hidden by the overlay. The same method was used to collect target location data on the physical model. After collecting both target and ground truth points in the NDI frame, target registration error (TRE) was
\begin{equation}
\mathrm{TRE}_k=\|x^{\mathrm{AR}}_k-x^{\mathrm{GT}}_k\|_2.
\end{equation}
The task was repeated with the full mandible exposed, teeth plus surrounding surface exposed, and teeth only exposed, giving 30 TRE values per exposure condition. For latency, a rigid motion onset triggered a timer, and the endpoint was the first rendered pose that reached 90\% of the final displacement. We conducted 30 motion trials.

\section{RESULTS}
\subsection{Target registration error}
Figure~\ref{fig:results}(a) shows the TRE distributions. Median TRE was
4.05~mm (interquartile range, IQR: 3.17--4.31~mm) with full exposure,
6.10~mm (IQR: 5.20--7.59~mm) with teeth and nearby surface exposed, and
7.10~mm (IQR: 5.82--8.67~mm) with teeth only. The teeth-only condition had
the widest observed range, 3.32--12.05~mm. The median across all 90 target
observations was 5.45~mm (IQR: 4.23--7.20~mm), as shown in
Fig.~\ref{fig:results}(b).

Treating the 30 anatomical target locations as matched repeated units, the Friedman test showed an overall
effect of exposure condition on TRE,
\(\chi^2_{\mathrm{F}}(2)=40.20\), \(N=30\),
\(p=1.87\times10^{-9}\), with Kendall's \(W=0.670\). Post-hoc paired
Wilcoxon signed-rank tests were Holm-adjusted across the three comparisons.

Relative to full exposure, the teeth-and-nearby-surface condition increased
TRE by a median within-target difference of 2.49~mm (IQR of paired
differences: 1.47--3.87~mm;
\(W_{\min}=6\), \(p_{\mathrm{Holm}}=5.22\times10^{-8}\),
\(r_{\mathrm{rb}}=0.974\)). The teeth-only condition increased TRE relative to full exposure by a median
within-target difference of 3.43~mm (IQR: 2.10--4.49~mm;
\(W_{\min}=3\), \(p_{\mathrm{Holm}}=2.79\times10^{-8}\),
\(r_{\mathrm{rb}}=0.987\)). The median paired increase from teeth and nearby surface to teeth only was
0.87~mm (IQR: 0.19--2.07~mm). Although teeth and nearby surface produced
lower TRE at 23 of the 30 matched targets, this contrast did not reach the
two-sided 0.05 threshold after correction for the three pairwise comparisons
(\(W_{\min}=139\), \(p_{\mathrm{Holm}}=0.0549\),
\(r_{\mathrm{rb}}=0.402\)). 

\subsection{Latency}
\begin{figure}[t]
\centering
\includegraphics[width=0.98\textwidth]{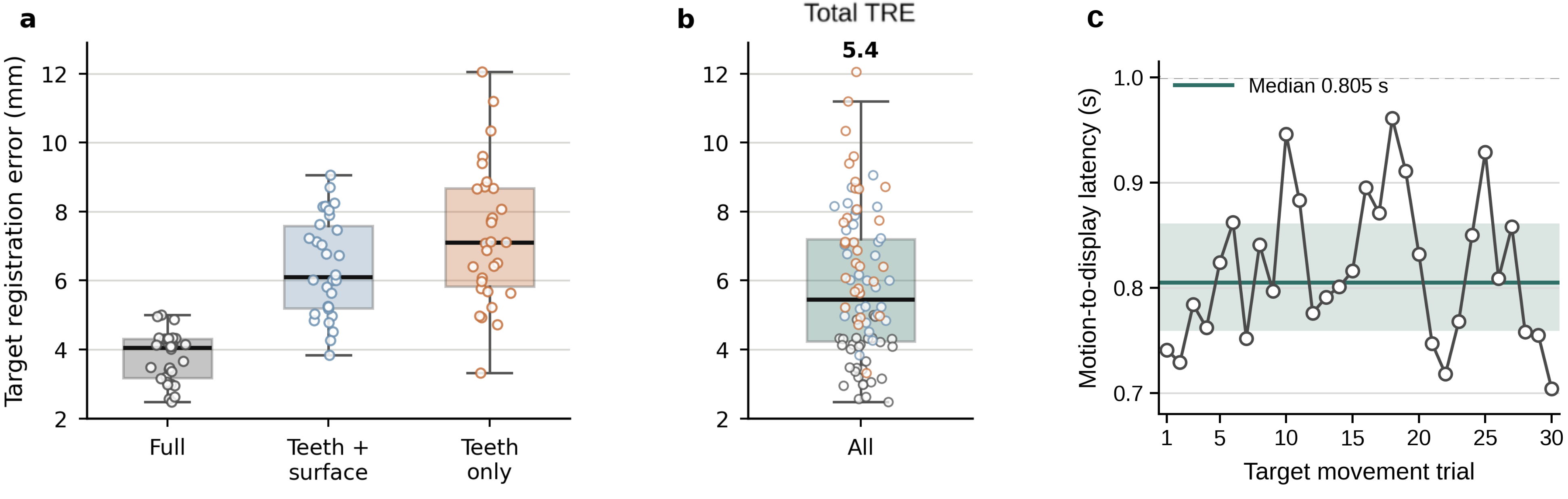}
\caption{TRE and latency data. (a) TRE under three exposure conditions. (b) All TRE. (c) Motion-to-display latency for 30 trials.}
\label{fig:results}
\end{figure}

Across 30 motion trials, median motion-to-display latency was 0.805~s (IQR: 0.759--0.861~s), and the maximum was 0.961~s. Every trial remained below 1~s. Figure~\ref{fig:results}(c)
reports the individual trial values.

\section{DISCUSSION AND CONCLUSION}
The proposed method targets a clinically important registration geometry, where a transoral depth image contains a small, irregular dental surface and may omit most of the CT mandible. The global stage gives high-curvature dental regions more influence, while the local stage matches observed points to the complete model without treating unobserved anatomy as an error. Equation~(\ref{eq:chain}) then transfers the preplanned guide independently of the mandible registration. This separation allows the surgeon to alter or replace the virtual guide without manufacturing another physical device. Since the current standard of using a printed cutting guide results in around 4~mm error \cite{weijs2016accuracy}, our TRE and latency results demonstrate the feasibility of replacing physical cutting guides with accurate, intuitive AR guidance.

Several limitations exist. Blood, saliva, or instruments can further occlude target depth points or create outliers, although teeth generally remain visible in a transoral approach. However, an extraoral or transcutaneous approach often exposes fewer teeth and therefore poses greater challenges. Visualization also needs more research. Rendering the full cutting guide and mandible may improve surgeon confidence but can occlude the operative field. Rendering only cut lines reduces occlusion but may provide too little context. Finally, interpolation improves visual smoothness without removing the measured latency. 

In conclusion, this work defines a markerless, teeth-aware, partial-to-whole registration pipeline for placing a virtual preoperative cutting guide on the mandible. It removes mounted fiducials and manual landmarks and updates the overlay after target motion. This work moves mandible reconstruction surgery, especially through the transoral approach, toward a more adaptable and accessible future.

\section{ACKNOWLEDGMENTS}       
This work was supported by the NIH under grant R01EB037685.

\clearpage
\bibliography{references}
\bibliographystyle{spiebib}

\end{document}